\documentclass[runningheads]{llncs}
\usepackage[T1]{fontenc}
\usepackage{graphicx,verbatim}
\usepackage{amssymb}
\usepackage{amsmath}
\usepackage{booktabs}
\usepackage[table]{xcolor}
\usepackage{wrapfig}
\usepackage[colorlinks=true, linkcolor=black, citecolor=blue, urlcolor=blue]{hyperref}

\usepackage{tikz,xcolor,hyperref}

\definecolor{lime}{HTML}{A6CE39}
\DeclareRobustCommand{\orcidicon}{
	\begin{tikzpicture}
	\draw[lime, fill=lime] (0,0) 
	circle [radius=0.16] 
	node[white] {{\fontfamily{qag}\selectfont \tiny ID}};
	\draw[white, fill=white] (-0.0625,0.095) 
	circle [radius=0.007];
	\end{tikzpicture}
	\hspace{-2mm}
}

\foreach \x in {A, ..., Z}{\expandafter\xdef\csname orcid\x\endcsname{\noexpand\href{https://orcid.org/\csname orcidauthor\x\endcsname}
			{\noexpand\orcidicon}}
}
			
\usepackage{tikz}
\usepackage{xparse}   

\DeclareRobustCommand{\authorpic}[2][5mm]{%
  \tikz[baseline={([yshift=-.25ex]current bounding box.center)}]{%
    \clip (0,0) circle (#1);
    \pgfmathsetlengthmacro{\picside}{sqrt(2)*#1}%
    \node at (0,0) {\includegraphics[width=\picside,height=\picside,keepaspectratio]{#2}};
    \draw[line width=0.4pt, color=white] (0,0) circle (#1);
  }%
}

\NewDocumentCommand{\AuthorWithPic}{O{5.5mm} O{0.20em} m m}{%
  \texorpdfstring{\authorpic[#1]{#4}\kern #2}{}%
  #3%
}

\usepackage{xcolor}
\definecolor{linkpinkix}{HTML}{EA335A} 

\definecolor{linkpink}{HTML}{EA335A}

\newcommand{\shadedlink}[2]{%
  \tikz[baseline=(n.base)]\node[
    fill=linkpink,
    fill opacity=0.5,
    text opacity=1,
    rounded corners=.3ex,
    inner xsep=.35em,
    inner ysep=.15em
  ] (n) {\href{#1}{\textcolor{blue!70!black}{#2}}};%
}

\begin{document}
\title{GraphTreeGen: Subtree-Centric Approach to Efficient and Supervised Graph Generation}

\author{%
  \AuthorWithPic[6mm][0.18em]{Yitong Luo\orcidA{}}{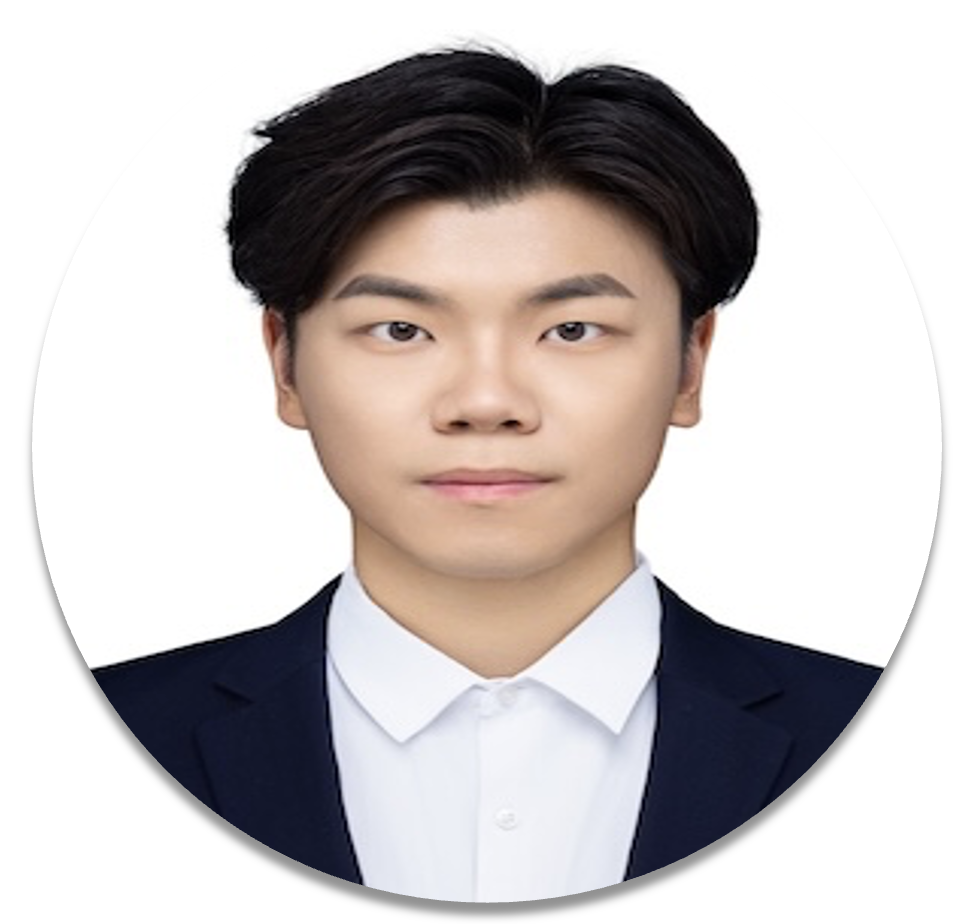} \and
  \AuthorWithPic[6mm][0.18em]{Islem Rekik\orcidA{}}{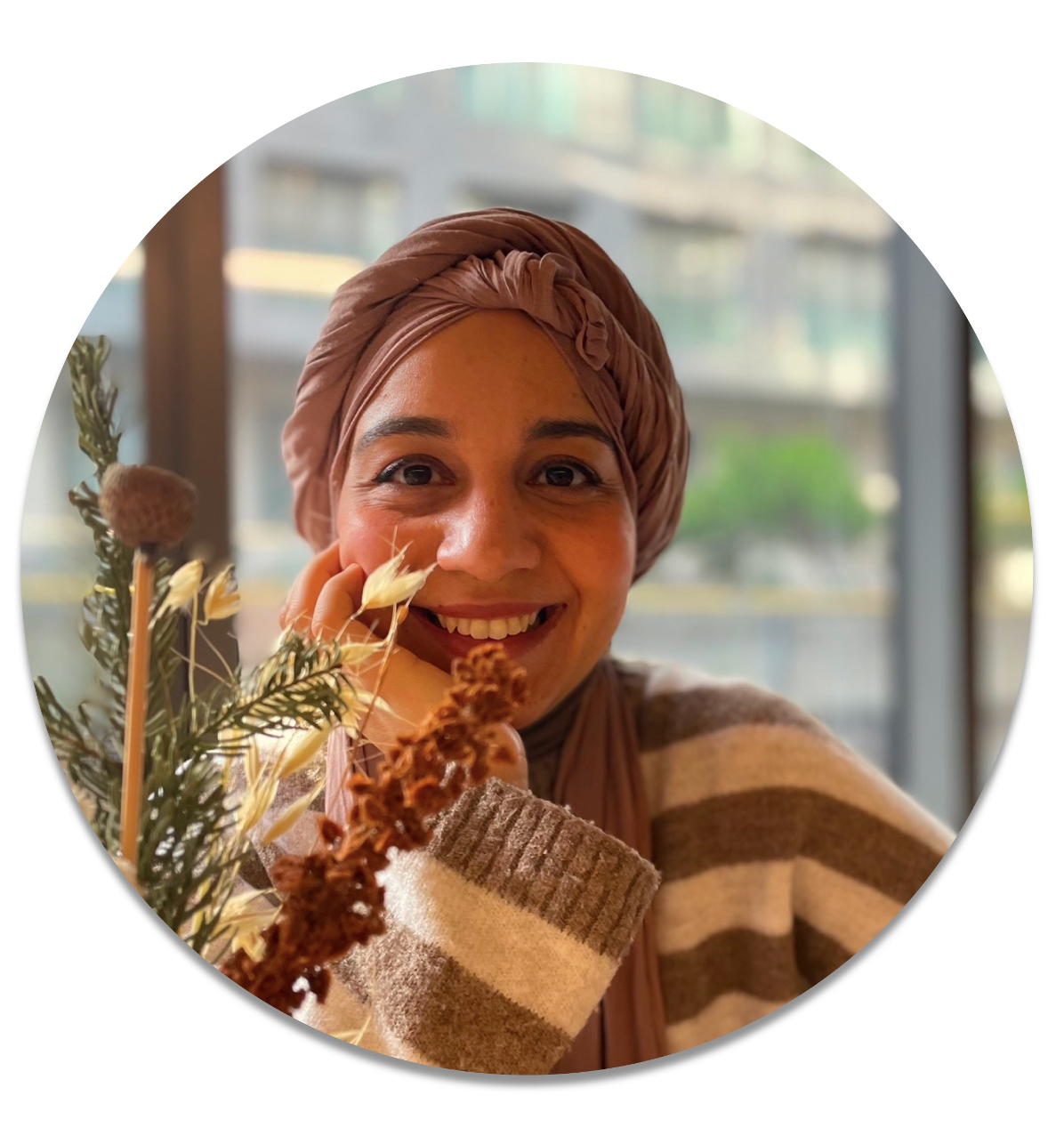}\thanks{Corresponding author: \email{i.rekik@imperial.ac.uk}, \url{http://basira-lab.com}, GitHub: \url{https://github.com/basiralab/GTG}}%
}

\authorrunning{Y. Luo and I. Rekik}
\institute{BASIRA Lab, Imperial-X and Department of Computing, Imperial College London, \\ London, UK \\
\email{yitong.luo24@imperial.ac.uk,i.rekik@imperial.ac.uk}\\
}

\maketitle              
\begin{abstract}

Brain connectomes, which represent neural connectivity as graphs, are crucial for understanding brain organization but are costly and time-consuming to acquire, motivating the development of generative methods. Recent progress in graph generative modeling offers a convenient data-driven alternative, enabling synthetic generation of connectomes and alleviating the dependence on extensive neuroimaging acquisitions. However, existing generative models still have the following key limitations: (i) they compress the entire graph into a single latent code, such as variational graph autoencoders (VGAEs), thereby blurring fine-grained local motifs; (ii) they rely on rich node attributes that brain connectomes typically do not provide, which diminishes reconstruction quality; (iii) edge-centric models like dual-graph frameworks emphasize topology but often overlook the accurate prediction of edge weights, sacrificing quantitative fidelity; and (iv) many state-of-the-art architectures, such as those using edge-conditioned convolutions, employ computationally expensive architectures, which results in significant memory demands and limits their scalability to larger connectomes.
To address these limitations, we propose \emph{\textbf{GraphTreeGen (GTG)}}, a novel subtree-centric generative framework explicitly designed for efficient and accurate brain connectome generation. GTG decomposes each connectome into entropy-guided $k$-hop trees that capture informative local structure, which are then encoded by a shared GCN. A bipartite message-passing layer merges subtree embeddings with global node features, and a dual-branch decoder jointly predicts edge existence and weights to rebuild the full adjacency matrix. Experiments demonstrate that GTG significantly outperforms state-of-the-art baselines in self-supervised task, and remains competitive in supervised settings, delivering higher structural fidelity and more precise edge weights while using far less memory. Its modular, resource-efficient design also lays the groundwork for extensions to connectome super-resolution and cross-modality synthesis. Our code is available at \url{https://github.com/basiralab/GTG/}.\footnote{This paper has been selected for a \textbf{Poster Presentation} at the PRIME MICCAI 2025 workshop. \shadedlink{https://youtu.be/EEcaKc0oN98}{[GTG YouTube Video]}.}


\end{abstract}
\section{Introduction}
The topological organization of brain graphs has been increasingly recognized as a key biomarker for detecting and characterizing neurodegenerative disorders, particularly Alzheimer’s disease (AD) and Parkinson’s disease (PD) \cite{pereira2015aberrant,pereira2016disrupted}. These conditions are commonly linked to abnormal patterns of neural connectivity, which can be systematically captured and analyzed through brain connectomes. Such disorders are often associated with disrupted patterns of neural connectivity, which can be effectively modeled using brain connectomes.
Leveraging such graph-based representations of brain organization facilitates a deeper understanding of disease mechanisms and supports the development of effective diagnostic tools \cite{parisot2018disease}. Brain connectomes represent the brain's network of neural connections as graphs, where nodes correspond to anatomical regions connected by weighted edges according to different modalities, such as morphological, functional and structural graphs \cite{bessadok2022graph,sporns2018graph}. However, obtaining comprehensive connectomes is challenging: brain imaging is costly and time-consuming, resulting in limited connectomic data availability \cite{sporns2011human,lichtman2014big}. Moreover, unlike many machine learning graphs, connectome nodes lack intrinsic feature vectors \cite{kim2021learning,isallari2020gsr,said2023neurograph}. 

In this context, there is a pressing need for effective generative methods capable of reconstructing and generating brain graphs by leveraging limited, noisy connectomes. The goal of these generative tasks is to accurately model missing or corrupted connectivity patterns, enabling the synthesis of realistic brain graphs that are useful for downstream analysis such as disease classification, cognitive prediction, or other neuroscientific tasks \cite{bessadok2021brain,xiao2024dyngnn,zong2023brainnetdiff}. Generative models provide valuable tools for overcoming data scarcity, enhancing the quality and fidelity of brain connectivity representations, and facilitating robust downstream applications, particularly in cases where additional data acquisition is prohibitively expensive or impractical. By learning comprehensive and meaningful representations directly from graph structures, these generative approaches enable more accurate and interpretable analyses of neural organization and function \cite{bessadok2022graph,isallari2020gsr,mhiri2021non,singh2024strongly}. Notably, previous work on graph super-resolution offers related but distinct solutions. GSR-Net \cite{isallari2020gsr}, the first framework to super-resolve a low-resolution brain connectome into a high-resolution one to emulate costly high-detail scans. Subsequent models like IMANGraphNet \cite{mhiri2021non}, uses computationally heavy edge-conditioned convolutions and runs into memory bottlenecks on full-size connectomes. Moreover, STP-GSR \cite{singh2024strongly} introduces a powerful generative framework that lifts node‑level representation into edge‑level learning via a primal‑dual graph transformation. By converting edges of the original brain graph into nodes in a dual graph, STP‑GSR enables GNNs to directly model connectivity relationships, preserving higher‑order topological structures. While effective at maintaining structural fidelity, STP‑GSR places greater emphasis on recovering the unweighted graph topology and relies on edge‑centering schemes that may underperform in learning precise connection strengths.

Considering graph generative task at the same resolution, for example graph reconstruction and same-node graph prediction, existing graph generative models face several limitations when applied to brain connectomes. Classical variational graph autoencoders (VGAEs) compress the entire graph into a global latent variable, creating a bottleneck that can obscure fine-grained topological features \cite{kipf2016variational}. Deep GNN encoders in such models also risk over-smoothing node representations – i.e. making embeddings of different regions overly similar – which is detrimental for preserving local connectivity structure. Standard graph autoencoders or graph convolution networks often assume the availability of rich node features; for example, the original VGAE formulation showed significantly improved performance when incorporating node attributes \cite{kipf2016variational,lin2023graph}. In connectomes, where nodes have no inherent features, such reliance becomes problematic \cite{isallari2020gsr}. On the other hand, generative adversarial approaches for graphs (e.g. GraphGAN \cite{wang2018graphgan}) and autoregressive models attempt to directly model the adjacency matrix distribution, but they may struggle to capture global properties of brain networks and can be unstable to train. Recently, diffusion-based generative models have emerged as a powerful paradigm in deep learning, including for graph data \cite{zong2023brainnetdiff,liu2023generative,rajadhyaksha2023diffusion}. While graph diffusion models can gradually refine a random graph toward realistic connectivity, they involve iterative sampling steps and have yet to be specifically tuned for brain connectomes. However, these models often rely on complex architectural components such as edge-conditioned convolutions, which result in high computational and memory demands—especially problematic for dense or large-scale brain graphs. Their training and inference pipelines are often resource-intensive and less suited to lightweight or scalable applications \cite{simonovsky2017dynamic}. This motivates the development of more memory-efficient and structurally aware generative models that are optimized for brain graph generation. Such models should balance architectural simplicity with topological fidelity, enabling effective representation learning while supporting broader neuroscientific applications.
\begin{figure}[t]
    \centering
    \includegraphics[width=1\linewidth]{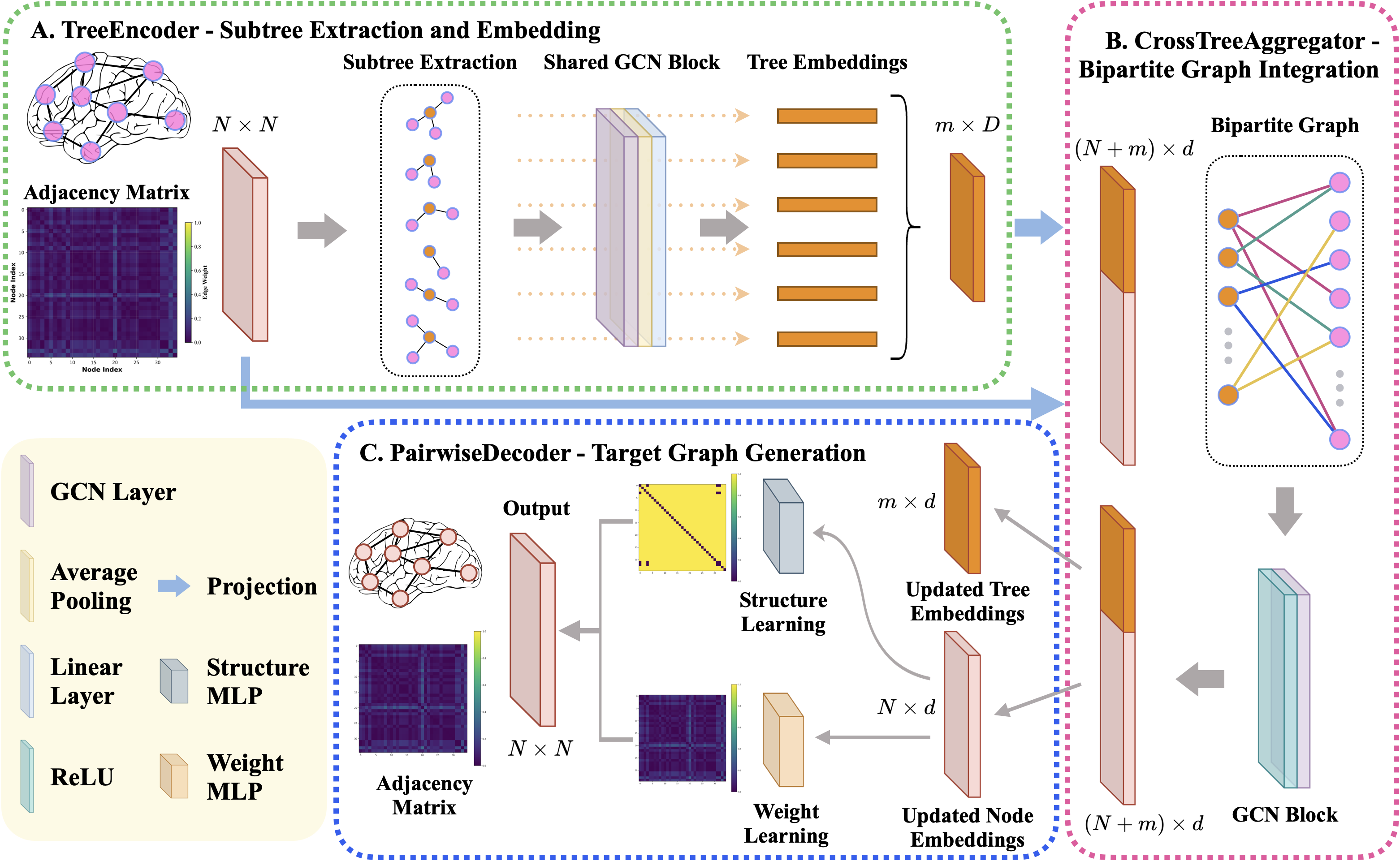}
    \caption{Illustration of the proposed GraphTreeGen (GTG) model. (A) TreeEncoder, which extracts subtrees and encodes them via a shared GCN into fixed-size embeddings; (B) CrossTreeAggregator, which builds a bipartite graph between nodes and subtrees to exchange information through GCN-based message passing; and (C) PairwiseDecoder, which reconstructs the adjacency matrix using parallel MLPs for structure and weight prediction based on the updated node and subtree embeddings.}
    \vspace{-20pt}
    \label{fig:GTG}
\end{figure}

To overcome the limitations of existing graph generative models, such as bottlenecks from global encoding, inefficiency on dense graphs, reliance on node attributes, and poor local structure preservation, we propose a novel and memory-efficient generative framework: \textbf{\textit{GraphTreeGen (GTG)}}. Instead of encoding the entire graph at once, GTG decomposes brain connectomes into multiple subtree-structured regions, enabling flexible modeling of local connectivity without over-compressing structural information. By applying a shared graph convolutional network (GCN) to these small subgraphs, GTG achieves significant memory savings and improved scalability, especially for large and dense connectomes that challenge traditional approaches like VGAE or IMANGraphNet. Furthermore, GTG eliminates the need for explicit node features by learning structural representations purely from edge patterns, which is an essential advantage for featureless brain graphs. Empirically, GTG consistently outperforms variational, adversarial, and dual-graph baselines in both structural fidelity and edge weight prediction. Its modular architecture also supports future extensions such as super-resolution and cross-modality synthesis.

GTG follows a three-stage design for brain connectome generation (\textbf{Fig.}~\ref{fig:GTG}), inspired by the junction tree variational autoencoder (JT-VAE) for molecular graphs \cite{jin2018junction} and the Bonsai gradient-free graph condensation framework \cite{gupta2024bonsai}. First, the model extracts a collection of rooted tree-structured subgraphs from the input connectome and computes initial embeddings for each root node. Then, it constructs a bipartite graph to enable message passing between tree roots and their corresponding nodes within the full graph, enhancing both node features and subtree-level embeddings. Finally, these enriched node representations are passed into a dual-branch pairwise decoder that reconstructs the full adjacency matrix, capturing both fine-grained local patterns and global brain connectivity. This design ensures GTG can efficiently learn biologically meaningful structure while remaining scalable to large, featureless brain graphs.

\section{Methodology}
GTG is designed to capture localized connectivity structures while ensuring global coherence, and consists of three main components, shown in \textbf{Fig.}~\ref{fig:GTG}: (1) a TreeEncoder that embeds local tree subgraphs (\textbf{Fig.}~\ref{fig:GTG}--A); (2) a CrossTreeAggregator that integrates local and global representations via bipartite message passing (\textbf{Fig.}~\ref{fig:GTG}--B); and (3) a PairwiseDecoder that reconstructs the full adjacency matrix (\textbf{Fig.}~\ref{fig:GTG}--C). The entire pipeline is trained end-to-end for graph generation, supporting both self-supervised and supervised learning.

\textbf{Problem statement.}
Let $\mathcal{G}^{(s)} = (\mathcal{V}^{(s)}, \mathcal{E}^{(s)})$ and $\mathcal{G}^{(t)} = (\mathcal{V}^{(t)}, \mathcal{E}^{(t)})$ denote the source and target undirected, weighted brain graphs, respectively. Each graph has its own node set, with $\mathcal{V}^{(s)} = \{v_1^{(s)}, \dots, v_N^{(s)}\}$ and $\mathcal{V}^{(t)} = \{v_1^{(t)}, \dots, v_N^{(t)}\}$, where $|\mathcal{V}^{(s)}| = |\mathcal{V}^{(t)}| = N$, and edge sets $\mathcal{E}^{(s)}$ and $\mathcal{E}^{(t)}$ capturing connectivity strengths. These graphs are represented by adjacency matrices $\mathbf{A}^{(s)}, \mathbf{A}^{(t)} \in \mathbb{R}^{N \times N}$, where $A_{ij}^{(s)}, A_{ij}^{(t)} \in \mathbb{R}_{\geq 0}$ encode the connection weights between nodes $v_i$ and $v_j$ in the source and target graphs, respectively. The node feature matrix $\mathbf{X}$ is defined as the adjacency matrix itself, with each row of the adjacency matrix serving as the feature vector of the corresponding node.
Given the source adjacency matrix $\mathbf{A}^{(s)}$, our goal is to learn a function $f_\theta$ that reconstructs the target graph $\mathbf{A}^{(t)}$ by leveraging a set of rooted tree-structured subgraphs $\{\mathcal{T}_1^{(s)}, \dots, \mathcal{T}_m^{(s)}\}$ extracted from $\mathcal{G}^{(s)}$. Each $\mathcal{T}_k^{(s)} \subseteq \mathcal{G}^{(s)}$ is a local subtree centered at a node $v_k^{(s)}$. The learning objective is to produce a predicted adjacency matrix $\widehat{\mathbf{A}}^{(t)}$ that closely approximates the ground-truth target:
\begin{equation}
    f_\theta\left( \{\mathcal{T}_1^{(s)}, \dots, \mathcal{T}_m^{(s)}\} \right) \rightarrow \widehat{\mathbf{A}}^{(t)} \approx \mathbf{A}^{(t)}.
\end{equation}
This formulation enables the model to capture localized structural information from the source graph while generating globally coherent predictions in the target graph space, thus supporting supervised brain graph generation tasks.

\textbf{TreeEncoder -- Subtree extraction and embedding (\textbf{Fig.}~\ref{fig:GTG}--A).}
To mitigate the effects of sparsity and noise in brain connectomes, we decompose each graph into a set of local $k$-hop tree-structured subgraphs ${ \mathcal{T}_1, \ldots, \mathcal{T}_m }$, each rooted at a selected node. The root nodes are selected based on their node entropy—the Shannon entropy of the degree‑weighted distribution over their 1‑hop-network. These subtrees preserve salient local structural patterns, such as motifs, clusters, or hub configurations, while reducing interference from long-range noisy edges. Each subtree $\mathcal{T}_k = (\mathcal{V}_k, \mathcal{E}_k)$ is encoded independently via a shared $L$-layer Graph Convolutional Network (GCN) that takes the subtree’s adjacency matrix $\widehat{\mathbf{A}}_k$ and node feature matrix $\mathbf{X}_k$ as input.

Following the GCN, we apply global average pooling to obtain a fixed-size representation of each subtree, and map it into a common latent space via a linear projection. The resulting embedding for subtree $\mathcal{T}_k$ is computed as:
\begin{equation}
    \mathbf{z}_k=\mathbf{W} \cdot \operatorname{AvgPool}\left(\operatorname{GCN}^{(L)}\left(\widehat{\mathbf{A}}_k, \mathbf{X}_k\right)\right)+\mathbf{b}
\end{equation}
where $\mathbf{W} \in \mathbb{R}^{D \times d}$ and $\mathbf{b} \in \mathbb{R}^{D}$ are learnable parameters, and $d$ is the GCN’s hidden dimension. Collectively, the encoder produces the matrix of subtree embeddings:
\begin{equation}
    \mathbf{T}=\left[\mathbf{z}_1 ; \mathbf{z}_2 ; \ldots ; \mathbf{z}_m\right] \in \mathbb{R}^{m \times D}
\end{equation}

This subtree-based encoding strategy allows the model to distill high-signal, low-noise structural information into compact latent representations, while supporting efficient, parallelizable computation.



\textbf{CrossTreeAggregator -- Bipartite graph integration (\textbf{Fig.}~\ref{fig:GTG}--B).}
While the TreeEncoder extracts local structural information from individual subtrees, it lacks global context and inter-subtree coordination. To address this, we introduce the CrossTreeAggregator module, which integrates subtree embeddings into a globally consistent representation by performing message passing over a bipartite graph.

Let $\mathcal{B} = (\mathcal{V}, \mathcal{T}, \mathcal{E}_B)$ denote a bipartite graph, where $\mathcal{V}$ is the set of original nodes in the brain graph, and $\mathcal{T} = { \mathcal{T}_1, \ldots, \mathcal{T}_m }$ is the set of subtrees extracted from $\mathcal{G}$. We construct $\mathcal{E}_B$ such that there is an edge between a node $v_i \in \mathcal{V}$ and a subtree $\mathcal{T}_k \in \mathcal{T}$ if and only if $v_i \in \mathcal{T}_k$.
Each node $v_i \in \mathcal{V}$ is initialized with a feature vector $\mathbf{x}_i \in \mathbb{R}^F$, and each subtree $\mathcal{T}_k$ is associated with an embedding $\mathbf{z}_k \in \mathbb{R}^D$ obtained from the TreeEncoder. We first project both node and subtree features into a shared latent space of dimension $d$, enabling unified message passing across the bipartite graph. All transformed features are stacked into a matrix $\mathbf{H}^{(0)} \in \mathbb{R}^{(N + m) \times d}$, where the first $N$ rows correspond to nodes and the remaining $m$ rows to subtrees.

We then apply $L$ layers of graph convolution over the bipartite graph:
\begin{equation}
    \mathbf{H}^{(\ell+1)}=\operatorname{ReLU}\left(\operatorname{GCN}^{(\ell)}\left(\mathcal{B}, \mathbf{H}^{(\ell)}\right)\right), \quad \ell=0,1, \ldots, L-1
\end{equation}
the updated embeddings are given by $\mathbf{H}_{node} = \mathbf{H}^{(L)}_{1:N}$ and $\mathbf{H}_{tree} = \mathbf{H}^{(L)}_{N+1:N+m}$.




\textbf{PairwiseDecoder -- Target graph generation (\textbf{Fig.}~\ref{fig:GTG}--C).}
After getting the updated node embeddings, we introduce a \emph{PairwiseDecoder} that predicts the existence and strength of connections between every pair of nodes. Given the updated node embeddings $\mathbf{H} = [\mathbf{h}_1; \mathbf{h}_2; \dots; \mathbf{h}_N] \in \mathbb{R}^{N \times d}$ from the \textit{CrossTreeAggregator}, the decoder predicts a symmetric adjacency matrix by evaluating pairwise interactions.

For each node pair $(i, j)$ with $i < j$, we construct a combined representation $\boldsymbol{\phi}_{ij} = [\mathbf{h}_i , | , \mathbf{h}_j , | , |\mathbf{h}_i - \mathbf{h}_j|] \in \mathbb{R}^{3d}$, where $|$ denotes concatenation, $|\cdot|$ denotes elementwise absolute difference, $\mathbf{h}_i$ and $\mathbf{h}_j$ represent each node’s individual context, while $|\mathbf{h}_i-\mathbf{h}_j|$ explicitly encodes how similar or dissimilar they are. Concatenating all three delivers both node-specific and relational signals in a unified feature, yielding more informative input for edge prediction.
This representation is passed through two parallel MLP branches: 
(1) structure branch $\operatorname{MLP}_{\text{struct}}: \mathbb{R}^{3d} \rightarrow \mathbb{R}$ computes the edge existence logit $\ell_{ij}$, outputs a symmetric matrix of edge logits $\boldsymbol{\ell} \in \mathbb{R}^{N \times N}$, where $\ell_{ij} = \ell_{ji}$ and $\ell_{ii} = -\infty$ to avoid self-loops.
(2) weight branch $\operatorname{MLP}_{\text{weight}}: \mathbb{R}^{3d} \rightarrow \mathbb{R}$ estimates the edge weight $\hat{A}^{(t)}_{ij}$, outputs a symmetric matrix of edge weights $\widehat{\mathbf{A}}^{(t)} \in \mathbb{R}^{N \times N}$, where $\widehat{A}^{(t)}_{ij} = \widehat{A}^{(t)}_{ji}$ and $\widehat{A}^{(t)}_{ii} = 0$.

\textbf{Topology and weight preserving loss function.}
To train the TreeGen model end-to-end, we define a composite loss function that jointly optimizes graph topology generation and edge weight regression. Given the predicted adjacency logits $\boldsymbol{\ell} \in \mathbb{R}^{N \times N}$ and edge weights $\widehat{\mathbf{A}}^{(t)} \in \mathbb{R}^{N \times N}$ from the \emph{PairwiseDecoder}, we supervise the model using two ground truth matrices: a binary adjacency matrix $\mathbf{A}^{(t)}_{logits} \in {0,1}^{N \times N}$ indicating edge existence, and a real-valued matrix $\mathbf{A}^{(t)} \in \mathbb{R}^{N \times N}$ encoding edge strengths.

The total training loss is defined as a weighted sum of two components:
\begin{equation}
    \mathcal{L}_{\text {total }}=\alpha \cdot \mathcal{L}_{\text {struct }}+\beta \cdot \mathcal{L}_{\text {weight }}
\end{equation}
where $\mathcal{L}_{\text{struct}}$ is the binary cross-entropy loss applied to edge existence logits, and $\mathcal{L}_{\text{weight}}$ is the mean absolute error between predicted and ground truth edge weights. In practice, we exploit the symmetry of undirected graphs by computing both losses only over the upper triangular portion of the matrices (excluding self-loops), and mirroring the values to complete the full prediction. This design ensures computational efficiency and respects the inherent structure of connectomes.



\section{Experiments}
\textbf{Evaluation dataset.} We evaluate our model on 341 brain connectomes from the ABIDE I dataset \cite{di2014autism}, comprising 186 neurotypical controls (NC) and 155 individuals with autism spectrum disorder (ASD). Each connectome is a $35 \times 35$ symmetric adjacency matrix without self-loops, derived from pairwise absolute differences of cortical morphology across 35 ROIs defined by the Desikan-Killiany atlas \cite{desikan2006automated}. Cortical features are extracted from T1-weighted MRI using the FreeSurfer pipeline \cite{fischl2012freesurfer}.
We use the ASD group for graph generation tasks. In the self-supervised setting, both source and target graphs represent the same cortical morphological measure; in the supervised setting, they correspond to two different cortical morphological measures. The dataset is split into five folds, with four used for cross-validation and one held out for final testing. We also evaluate classification performance between NC and ASD connectomes to assess the quality of generated representations.

\noindent \textbf{Parameter setting.} Each graph has 35 nodes with adjacency-based node features. The TreeEncoder and CrossTreeAggregator use hidden and output dimensions of 32 and 16, respectively, with 2 layers and 15 subtrees ($m=15$, $k=1$, which we tested to be the best number of subtrees extracted from the dataset). The PairwiseDecoder uses MLPs with hidden size 32. Models are trained for 50 epochs using Adam (learning rate $10^{-3}$, batch size 8). Loss weights are set to $\alpha=10.0$ (structure), $\beta=5$ (weight), and seed 42 ensures reproducibility. All experimental parameters are thoroughly validated through multiple experiments, and the optimal combination is selected.

\noindent \textbf{Evaluation and comparison methods.}
We compare GTG with three representative state-of-the-art (SOTA) methods: IMANGraphNet \cite{mhiri2021non}, which performs inter-modality brain graph synthesis via adversarial alignment; STP-GSR \cite{singh2024strongly}, which improves topological fidelity through dual-graph super-resolution; and VGAE \cite{kipf2016variational}, a classic variational autoencoder that reconstructs graphs using GCN-based latent embeddings. The reconstructed graphs are evaluated using ten comprehensive metrics spanning edge-level accuracy, node centrality preservation, and global structural consistency. Specifically: 
(1) \emph{MAE} measures the mean absolute error of predicted edge weights; 
(2) \emph{$\text{MAE}_{\text{deg}}$} evaluates degree centrality error, reflecting node connectivity preservation; 
(3) \emph{$\text{MAE}_{\text{bc}}$} captures changes in betweenness centrality, related to shortest-path roles; 
(4) \emph{$\text{MAE}_{\text{ec}}$} assesses eigenvector centrality discrepancies based on influence propagation; 
(5) \emph{$\text{MAE}_{\text{ic}}$} measures information centrality mismatches in communication efficiency; 
(6) \emph{$\text{MAE}_{\text{pr}}$} quantifies PageRank score deviations; 
(7) \emph{$\text{MAE}_{\text{katz}}$} captures Katz centrality error from attenuated walks; 
(8) \emph{$\text{MAE}_{\text{lap}}$} measures Laplacian centrality variation under node removal; 
(9) \emph{Clustering Coefficient Difference} reflects shifts in local triangle density; 
and (10) \emph{Laplacian Frobenius Distance} quantifies global spectral divergence. 
These metrics collectively provide a multi-scale evaluation framework for graph generation, spanning local edges, node influence, and global spectral structure.

\emph{\textbf{Ablation analysis.}}
To assess the contribution of the pairwise decoder components in GTG, we conduct an ablation study by separately removing the structure prediction branch ($\operatorname{MLP}_{struct}$) and the weight prediction branch ($\operatorname{MLP}_{weight}$). As shown in \textbf{Table}~\ref{tab:ablation_study}, the full model outperforms both ablated variants across all metrics, confirming that jointly modeling edge existence and weight leads to more accurate and structurally faithful graph generation.

\begin{table}[h]
\vspace{-15pt}
\centering
\small
\caption{Ablation study of GTG.}
\vspace{-8pt}
\resizebox{\textwidth}{!}{%
\begin{tabular}{lcccccccccc}
\toprule
\textbf{Model} & \textbf{MAE} & \textbf{MAE$_{\deg}$} & \textbf{MAE$_{BC}$} & \textbf{MAE$_{EC}$} & \textbf{MAE$_{IC}$} & \textbf{MAE$_{PR}$} & \textbf{MAE$_{\mathrm{Katz}}$} & \textbf{MAE$_{\mathrm{Lap}}$} & \textbf{Clust.\ diff.} & \textbf{Lap.\ Fro.\ dist.} \\
\midrule
w/o $\operatorname{MLP}_{struct}$  & 0.4202 & 0.0106 & 0.0890 & 0.0349 & 0.2198 & 0.0067 & 0.0134 & 0.0262 & 0.7831 & 86.0644 \\
w/o $\operatorname{MLP}_{weight}$  & 0.0104 & 0.0106 & 0.0905 & 0.0046 & 0.0028 & 0.0009 & 0.0022 & 0.0037 & 0.0207 & 1.9666 \\
full             & \cellcolor{green!25}\textbf{0.0092} & \cellcolor{green!25}\textbf{0.0106} & \cellcolor{green!25}\textbf{0.0880} & \cellcolor{green!25}\textbf{0.0036} & \cellcolor{green!25}\textbf{0.0027} & \cellcolor{green!25}\textbf{0.0007} & \cellcolor{green!25}\textbf{0.0019} & \cellcolor{green!25}\textbf{0.0030} & \cellcolor{green!25}\textbf{0.0166} & \cellcolor{green!25}\textbf{1.7701} \\
\bottomrule
\end{tabular}%
}
\label{tab:ablation_study}
\vspace{-15pt}
\end{table}

\emph{\textbf{Comparison with SOTA methods.}}
We first compared our GTG model with several baselines on self-supervised learning task. As shown in \textbf{Table}~\ref{tab:comparison_metrics_selfsupervised}, 
\begin{table}[h]
\centering
\small
\caption{Self$-$supervised graph generation metrics (mean$_{\pm\mathrm{std}}$). Bold values with green background denote best performance.}
\vspace{-8pt}
\resizebox{\textwidth}{!}{%
\begin{tabular}{lcccccccccc}
\toprule
\textbf{Model}   & \textbf{MAE}                 & \textbf{MAE$_{\deg}$}        & \textbf{MAE$_{BC}$}        & \textbf{MAE$_{EC}$}        & \textbf{MAE$_{IC}$}        & \textbf{MAE$_{PR}$}        & \textbf{MAE$_{\mathrm{Katz}}$} & \textbf{MAE$_{\mathrm{Lap}}$} & \textbf{Clust.\ diff.}        & \textbf{Lap.\ Fro.\ dist.}      \\
\midrule
IMAN \cite{mhiri2021non}     & $0.8955_{\pm0.0296}$         & $0.0694_{\pm0.0000}$         & $0.0874_{\pm0.0059}$       & $0.0255_{\pm0.0065}$       & $0.3994_{\pm0.0768}$       & $0.0051_{\pm0.0006}$       & $0.0896_{\pm0.0055}$          & $0.0191_{\pm0.0030}$          & $0.0524_{\pm0.0186}$          & $191.3031_{\pm13.1412}$        \\
STP-GSR \cite{singh2024strongly}  & $0.2908_{\pm0.4095}$         & $0.2606_{\pm0.4269}$         & $0.3189_{\pm0.3933}$       & $0.2780_{\pm0.4169}$       & $0.2587_{\pm0.4280}$       & $0.2545_{\pm0.4304}$       & $0.2582_{\pm0.4283}$          & $0.2696_{\pm0.4217}$          & $0.3558_{\pm0.3719}$          & $  5.9569_{\pm 2.9069}$        \\
VGAE \cite{kipf2016variational}     & $0.4300_{\pm0.0020}$         & $0.0694_{\pm0.0000}$         & $0.1005_{\pm0.0015}$       & $0.0374_{\pm0.0004}$       & $0.2070_{\pm0.0011}$       & $0.0069_{\pm0.0000}$       & $0.0407_{\pm0.0016}$          & $0.0274_{\pm0.0002}$          & $0.1852_{\pm0.0044}$          & $ 84.9996_{\pm 0.4933}$        \\
\midrule
BonsaiGen & $0.0567_{\pm0.0004}$         & $0.0694_{\pm0.0000}$         & \cellcolor{green!20}$\mathbf{0.0840_{\pm0.0000}}$ & $0.0347_{\pm0.0000}$       & $0.0069_{\pm0.0004}$ & $0.0067_{\pm0.0000}$       & $0.0133_{\pm0.0000}$          & $0.0261_{\pm0.0000}$          & $0.7079_{\pm0.0117}$          & $  7.5163_{\pm 0.1928}$        \\
\textbf{GTG}   & \cellcolor{green!20}$\mathbf{0.0081_{\pm0.0008}}$ & \cellcolor{green!20}$\mathbf{0.0106_{\pm0.0000}}$ & $0.0890_{\pm0.0035}$       & \cellcolor{green!20}$\mathbf{0.0035_{\pm0.0006}}$ & \cellcolor{green!20}$\mathbf{0.0021_{\pm0.0004}}$ & \cellcolor{green!20}$\mathbf{0.0007_{\pm0.0001}}$ & \cellcolor{green!20}$\mathbf{0.0018_{\pm0.0002}}$   & \cellcolor{green!20}$\mathbf{0.0029_{\pm0.0005}}$  & \cellcolor{green!20}$\mathbf{0.0154_{\pm0.0021}}$  & \cellcolor{green!20}$\mathbf{1.5838_{\pm 0.1385}}$ \\
\bottomrule
\end{tabular}%
}
\label{tab:comparison_metrics_selfsupervised}
\end{table}
GTG outperforms all baselines across nearly all evaluation metrics, including edge weight accuracy, multiple centrality-based measures, and both local and global structural fidelity. While GTG slightly underperforms BonsaiGen in betweenness centrality, this may be attributed to BonsaiGen’s reverse KNN strategy, which better preserves bridge nodes in shortest-path structures. In contrast, GTG selects subtrees based on node entropy, emphasizing locally informative regions, which may underrepresent long-range connectivity. Despite this, GTG demonstrates strong generalization and structural awareness, achieving the most robust and consistent generative performance overall.

\begin{figure}[t]
    \centering
    \includegraphics[width=1\linewidth]{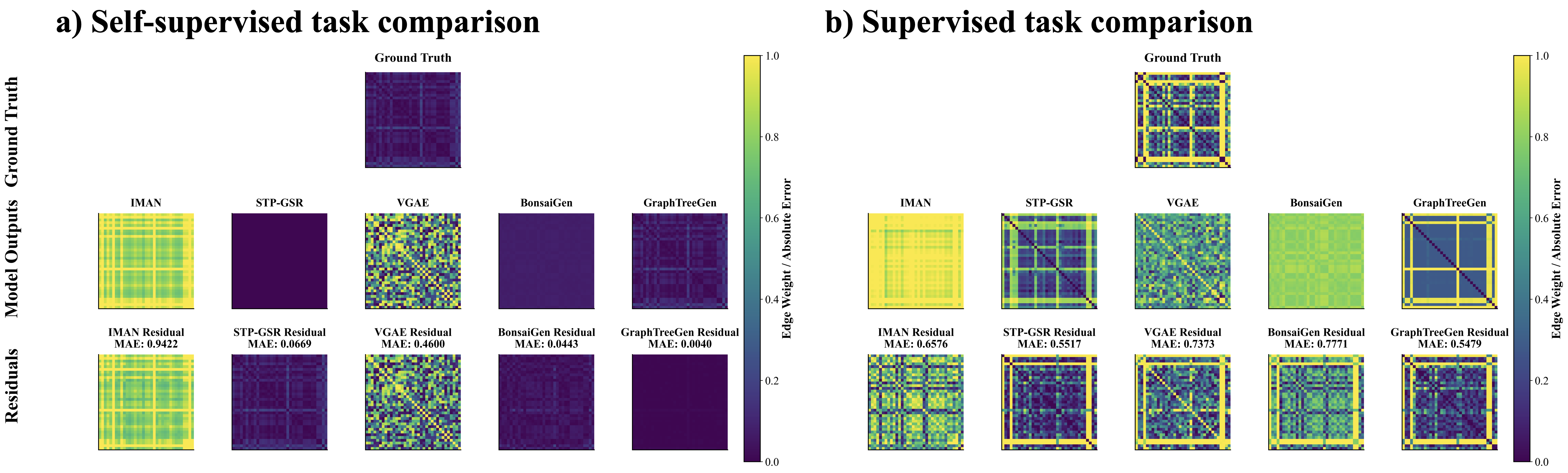}
    \caption{Illustration of reconstructed brain connectome adjacency matrix and the residual error comparing with ground truth adjacency matrix. a) Comparison under self-supervised task, and b) comparison under supervised task.}
    \vspace{-15pt}
    \label{fig:visualized_comparison}
\end{figure}

Moreover, as illustrated in \textbf{Fig.}~\ref{fig:visualized_comparison}-a, we visualize the generation outputs of each model on a representative sample under the self-supervised setting. Compared to IMAN, VGAE, STP-GSR, and BonsaiGen, our GraphTreeGen (GTG) exhibits the closest resemblance to the ground truth, both in terms of structural layout and edge weight distribution. The residual maps further confirm GTG’s superiority, with a significantly lower MAE of 0.0040, far outperforming all other baselines. These qualitative results complement our quantitative findings, reinforcing the strength of GTG in producing topologically faithful and weight-accurate brain graph generations.

In the supervised graph generation task, shown in \textbf{Table}~\ref{tab:comparison_metrics_supervised}, our proposed GTG model achieves the best performance in 4 out of 10 metrics and ranks second-best in nearly all others, demonstrating consistent and competitive generation capability. While models such as IMAN and VGAE slightly outperform GTG on a few specific centrality metrics, GTG still maintains top-tier accuracy overall, including the lowest MAE and degree centrality errors, which reflect strong structural preservation. 

\begin{table}[h]
\vspace{-15pt}
\centering
\small
\caption{Supervised graph generation metrics (mean$_{\pm\mathrm{std}}$). Dark green: best (lowest), light green: second best.}
\vspace{-8pt}
\resizebox{\textwidth}{!}{%
\begin{tabular}{lcccccccccc}
\toprule
\textbf{Model}   & \textbf{MAE}                 & \textbf{MAE$_{\deg}$}        & \textbf{MAE$_{BC}$}        & \textbf{MAE$_{EC}$}        & \textbf{MAE$_{IC}$}        & \textbf{MAE$_{PR}$}        & \textbf{MAE$_{\mathrm{Katz}}$} & \textbf{MAE$_{\mathrm{Lap}}$} & \textbf{Clust.\ diff.}        & \textbf{Lap.\ Fro.\ dist.}      \\
\midrule
IMAN \cite{mhiri2021non}     & $0.6658_{\pm0.0673}$         & $0.0694_{\pm0.0000}$         & $0.0933_{\pm0.0050}$       & \cellcolor{green!50}$\mathbf{0.0166_{\pm0.0076}}$       & $0.2784_{\pm0.0405}$       & \cellcolor{green!50}$\mathbf{0.0067_{\pm0.0005}}$ & $0.1103_{\pm0.0047}$          & \cellcolor{green!50}$\mathbf{0.0149_{\pm0.0033}}$  & \cellcolor{green!50}$\mathbf{0.0712_{\pm0.0117}}$  & $113.6312_{\pm29.8461}$        \\
STP-GSR \cite{singh2024strongly}  & \cellcolor{green!20}$\mathbf{0.5950_{\pm0.1451}}$         & $0.1805_{\pm0.2946}$         & $0.1887_{\pm0.1664}$       & $0.1347_{\pm0.1908}$       & $0.2434_{\pm0.2033}$       & $0.1112_{\pm0.1808}$         & $0.1585_{\pm0.1776}$          & $0.1242_{\pm0.1795}$          & $0.3419_{\pm0.1144}$          & \cellcolor{green!20}$\mathbf{110.0151_{\pm 5.5866}}$        \\
VGAE \cite{kipf2016variational}     & $0.6973_{\pm0.0011}$         & $0.0695_{\pm0.0000}$         & \cellcolor{green!50}$\mathbf{0.0887_{\pm0.0006}}$ & $0.0464_{\pm0.0002}$       & \cellcolor{green!50}$\mathbf{0.0787_{\pm0.0015}}$ & $0.0113_{\pm0.0000}$         & $0.0947_{\pm0.0003}$          & $0.0368_{\pm0.0001}$          & $0.3586_{\pm0.0048}$          & $131.5964_{\pm0.5870}$         \\
\midrule
BonsaiGen & $0.7559_{\pm0.0040}$         & \cellcolor{green!20}$0.0696_{\pm0.0000}$         & $0.0905_{\pm0.0000}$       & $0.0454_{\pm0.0000}$       & \cellcolor{green!20}$\mathbf{0.0923_{\pm0.0036}}$       & $0.0111_{\pm0.0000}$         & \cellcolor{green!20}$\mathbf{0.0940_{\pm0.0001}}$          & $0.0361_{\pm0.0000}$          & $0.7708_{\pm0.0200}$          & $117.0392_{\pm0.1071}$         \\
\textbf{GTG}   & \cellcolor{green!50}$\mathbf{0.5719_{\pm0.0060}}$ & \cellcolor{green!50}$\mathbf{0.0108_{\pm0.0000}}$ & \cellcolor{green!20}$\mathbf{0.0905_{\pm0.0000}}$       & \cellcolor{green!20}$\mathbf{0.0270_{\pm0.0018}}$ & $0.1000_{\pm0.0048}$       & \cellcolor{green!20}$\mathbf{0.0077_{\pm0.0003}}$         & \cellcolor{green!50}$\mathbf{0.0587_{\pm0.0040}}$  & \cellcolor{green!20}$\mathbf{0.0232_{\pm0.0014}}$  & \cellcolor{green!20}$\mathbf{0.3418_{\pm0.0142}}$          & \cellcolor{green!50}$\mathbf{109.4230_{\pm1.9924}}$ \\
\bottomrule
\end{tabular}%
}
\label{tab:comparison_metrics_supervised}
\vspace{-15pt}
\end{table}

Additionally, \textbf{Fig.}~\ref{fig:visualized_comparison}-b presents model outputs under the supervised learning task. While STP-GSR demonstrates slightly better structural fidelity, owing to its dual-graph topological emphasis, our GTG still achieves competitive performance in preserving structural patterns. Notably, GTG provides more accurate edge weight predictions than all other baselines except STP-GSR, as evidenced by lower residuals. 

\begin{wraptable}{r}{0.45\textwidth}
\vspace{-25pt}
\centering
\footnotesize
\caption{Training memory usage.}
\begin{tabular}{lr}
\toprule
\textbf{Model} & \textbf{Memory Usage (MB)} \\
\midrule
IMAN \cite{mhiri2021non}           & 570.54 \\
STP-GSR \cite{singh2024strongly}        & 512.48 \\
VGAE \cite{kipf2016variational}           & 344.59 \\
\midrule
BonsaiGen       &  \cellcolor{green!50}$31.60$ \\
GTG-10          & \cellcolor{green!20}\textbf{150.65} \\
GTG-15          & 161.22 \\
GTG-20          & 184.66 \\
\bottomrule
\end{tabular}
\label{tab:memory_comparison}
\vspace{-15pt}
\end{wraptable}

This balance between structure preservation and weight accuracy highlights GTG’s robustness, making it particularly suitable for practical applications where both topology and connectivity strength matter.
Importantly, GTG achieves this with significantly lower memory overhead, making it not only effective but also highly efficient and scalable, especially advantageous for large-scale or resource-constrained applications.

\emph{\textbf{Memory usage.}}
\textbf{Table}~\ref{tab:memory_comparison} shows that GTG significantly reduces memory usage compared to most of the baselines, except BonsaiGen. We adapted BonsaiGen by extending the core Bonsai graph condensation method, retaining its reverse KNN-based subtree selection, and added a decoder to reconstruct full graphs. While BonsaiGen achieves the lowest memory footprint, it performs notably worse than GTG in generation quality. The variants GTG-10/15/20 reflect the number of extracted subtrees, with memory usage increasing moderately as more subtrees are selected. Overall, GTG provides a compelling trade-off between accuracy and memory efficiency.

\emph{\textbf{Classification power analysis.}}
For the classification experiment, we first pre‑train GraphTreeGen in a label‑free manner on the combined ASD and NC graphs, then freeze the encoder and assess a downstream classifier on the resulting embeddings. As shown in \textbf{Table}~\ref{tab:classification_results}, GTG demonstrates strong performance, ranking second in accuracy and precision, closely following IMAN GraphNet, and achieving recall values comparable to BonsaiGen. Despite this slight gap, GTG offers a significantly lower memory footprint and superior graph generation quality, as demonstrated in earlier evaluations. These results highlight GTG's ability to generate structurally faithful representations that remain highly effective for downstream analysis, making it a competitive and resource-efficient alternative to more complex models.

\begin{table}[h]
\vspace{-15pt}
\centering
\small
\caption{Classification performance (mean$_{\pm\mathrm{std}}$). Dark green: best (highest), light green: second best.}
\vspace{-8pt}
\begin{tabular}{lccc}
\toprule
\textbf{Model} & \textbf{Accuracy} & \textbf{Precision} & \textbf{Recall} \\
\midrule
IMAN \cite{mhiri2021non}  & \cellcolor{green!50}$\mathbf{0.6014_{\pm0.0162}}$ & \cellcolor{green!50}$\mathbf{0.5810_{\pm0.0242}}$ & $0.4032_{\pm0.0361}$ \\
STP-GSR \cite{singh2024strongly}      & $0.5399_{\pm0.0314}$ & $0.3560_{\pm0.2228}$ & $0.1452_{\pm0.0869}$ \\
VGAE \cite{kipf2016variational}         & $0.5399_{\pm0.0415}$ & $0.4679_{\pm0.2337}$ & $0.0968_{\pm0.0395}$ \\
\midrule
BonsaiGen     & $0.5399_{\pm0.0188}$ & $0.4868_{\pm0.0221}$ & \cellcolor{green!50}$\mathbf{0.4597_{\pm0.0352}}$ \\
\textbf{GTG}       & \cellcolor{green!20}$0.5870_{\pm0.0126}$ & \cellcolor{green!20}$0.5540_{\pm0.0267}$ & \cellcolor{green!20}$0.4355_{\pm0.0361}$ \\
\bottomrule
\end{tabular}
\label{tab:classification_results}
\vspace{-15pt}
\end{table}

\section{Conclusion}
In this paper, we introduced \emph{\textbf{GraphTreeGen (GTG)}}, a novel and memory-efficient framework designed for brain connectome generation. Unlike traditional approaches that either compress global graph structure or rely heavily on computationally expensive edge modeling, GTG introduced a subtree-centric generation paradigm that effecively balanced local detail preservation and global structural generation.

Our method featured two core innovations: (1) a tree-aware encoder that captured localized structural patterns by decomposing the graph into entropy-guided subtrees and encoding them using a shared GCN, and (2) a cross-level bipartite aggregation strategy that enabled interactions between subtree embeddings and global node representations, bridging local and global semantics. This was followed by a dual-branch pairwise decoder that jointly predicted both edge existence and edge weights, thus ensuring structural fidelity and accurate connectivity reconstruction.

Extensive experiments on brain connectome datasets demonstrated that GraphTreeGen consistently outperformed state-of-the-art baselines in the self-supervised setting, achieving significant improvements across edge-level accuracy, centrality-based topology metrics, and spectral distance. In the supervised task, it achieved performance close to the best models while retaining superior scalability. Notably, our framework consumed less than one-third the memory of competitive baselines, making it highly suitable for large-scale connectomes or dense functional graphs where computational resources are limited. These results highlighted GTG as a practical, accurate, and extensible solution for structure-aware brain graph generation. The modular architecture also facilitates extensions to connectome super-resolution and cross-modality graph synthesis.

%
%
%
\bibliographystyle{splncs}
\bibliography{bibliography}
%




\end{document}